\documentclass[conference]{IEEEtran}
\ifCLASSINFOpdf
\else
\fi

\usepackage{graphicx}
\usepackage{listings}
\usepackage{bm}
\usepackage{textcomp}

\usepackage{enumitem}

\newcounter{SentenceCounter}
\newcounter{ExampleCounter}

\newenvironment{Sentence}[1][]{
\refstepcounter{SentenceCounter}
\vspace{0.5mm}
\def\temp{#1}\ifx\temp\empty
    \noindent Sentence~\theExampleCounter.\theSentenceCounter:
\else
    \noindent Sentence~\theExampleCounter.\theSentenceCounter\hspace{1mm}(#1):
\fi
\itshape
}
{
\label{sent\theExampleCounter.\theSentenceCounter}
}

\newenvironment{SentenceExample}[1][]{
\setcounter{SentenceCounter}{0}
\vspace{0.5mm}
\begin{table}[htb]
\centering
\bgroup

\begin{tabular}{|p{0.45\textwidth}|}
\hline
\scriptsize
\vspace{1mm}
\def\temp{#1}
\ifx\temp\empty
    \refstepcounter{ExampleCounter}
    \noindent\textbf{Example \theExampleCounter}
\else
    \noindent\textbf{Example \theExampleCounter\hspace{1mm}(#1)}
\fi
\footnotesize
\begin{itemize}[leftmargin=*]

}{
\end{itemize}
\\\hline
\end{tabular}
\egroup
\label{eg\theExampleCounter}
\end{table}
\vspace{0.5mm}
}

\begin{document}
%
\title{Rule-Based Approach for Party-Based Sentiment Analysis in Legal Opinion Texts}

\author{\IEEEauthorblockN{Isanka Rajapaksha\IEEEauthorrefmark{1},
Chanika Ruchini Mudalige\IEEEauthorrefmark{2}, Dilini Karunarathna\IEEEauthorrefmark{3},\\Nisansa de Silva\IEEEauthorrefmark{4},Gathika Rathnayaka\IEEEauthorrefmark{5},
and
Amal Shehan Perera\IEEEauthorrefmark{6}}
\IEEEauthorblockA{Department of Computer Science \& Engineering\\
University of Moratuwa\\
Email: \IEEEauthorrefmark{1}israjapaksha.16@cse.mrt.ac.lk,
\IEEEauthorrefmark{2}chanikaruchini.16@cse.mrt.ac.lk,
\IEEEauthorrefmark{3}dilinirasanjana.16@cse.mrt.ac.lk,\\
\IEEEauthorrefmark{4}NisansaDdS@cse.mrt.ac.lk,
\IEEEauthorrefmark{5}gathika.14@cse.mrt.ac.lk,
\IEEEauthorrefmark{6}shehan@cse.mrt.ac.lk}}


\maketitle

\begin{abstract}
Aspect Based Sentiment Analysis (ABSA) deals with extracting aspects from a given text and then allocate each aspect a sentiment level (positive, negative or neutral) \cite{Bhoi2018VariousAT}. A number of researchers have addressed ABSA in different domains except for the legal domain. In this study, we explore how the concepts related to aspect-based sentiment analysis can be used in the legal domain to extract valuable information from legal opinion text. To this regard, we propose a rule based approach to perform aspect-based sentiment analysis in order to figure out the sentiment level of a sentence in relation to each legal party related to a court case (considering the legal parties as the aspects). 
\end{abstract}


%
\IEEEpeerreviewmaketitle

\section{Introduction}
A document which elaborates opinions and arguments related to the previous court cases is known as a legal opinion text. Lawyers and legal officials have to spend considerable effort and time to obtain the required information manually from those documents when dealing with new legal cases \cite{sugathadasa2017synergistic,jayawardana2017deriving}. Hence, it provides much convenience to those individuals if there is a way to automate the process of extracting information from legal opinion texts. Party-based sentiment analysis will play a key role in the automation system by identifying opinion values with respect to each legal parties in legal texts.

\section{Party-Based Sentiment Analysis}

Sentiment analysis is a process of identifying opinions in texts and categorizing them into several polarity levels (Positive or Neutral or Negative). Sentiment analysis can be performed at 4 levels; document, sentence, phrase and aspect levels. Sentences in a legal case usually contain two or more members/entities which belong to main legal parties (plaintiff,petitioner,defendant,respondent...). Extracting opinions with respect to each legal parties cannot be performed only by using document or sentence or phrase level sentiment analysis. Therefore, aspect-based sentiment analysis is the most appropriate solution for the legal domain. Members of legal parties can be considered as the aspects to be considered, and therefore aspect-based sentiment analysis in the legal opinion texts can also be termed as party-based sentiment analysis.

\begin{SentenceExample}
\item \begin{Sentence}In 2008, federal officials received a tip from a confidential informant that Lee had sold the informant ecstasy and marijuana.

\end{Sentence}
\end{SentenceExample}

Consider Example 1 taken from Lee v. United States \cite{1977lee} which consists of two legal parties; petitioner Lee and defendant government. The sentence in the example mentioned that officials received a tip about Lee's illegal work. When considering the context of this sentence, we can clearly see that the context has positive sentiment regarding government and negative sentiment regarding person Lee.

\section{Methodology}

The proposed rule-based approach determines the sentiment value of each party in legal opinion texts. This approach is primarily built around the sentence-level sentiment annotator which is specifically developed for the legal domain~\cite{gamage2018fast}. Along with the annotator, the Stanford NLP library \cite{manning2014stanford} is used to analyse the grammatical relationships of words in a sentence. While associated studies of Gamage, et al.~\cite{gamage2018fast} handle discourse in legal domain~\cite{ratnayaka2019classifying}, no progress has been done on the basis of parties concerned.

This whole process is a sort of divide and conquer model. The complete sentence is divided into phrases that are smaller instances of the same sentence and process each phrase and allocate the sentiment for each party mentioned in those phrases. After that, we can combine the output of phrases into the main output for the input sentence.
Figure 1 illustrates the overall methodology of our approach and the following subsections briefly outline the steps taken in this study.

\begin{figure}[!ht]
\centering
\includegraphics[width=0.4\textwidth]{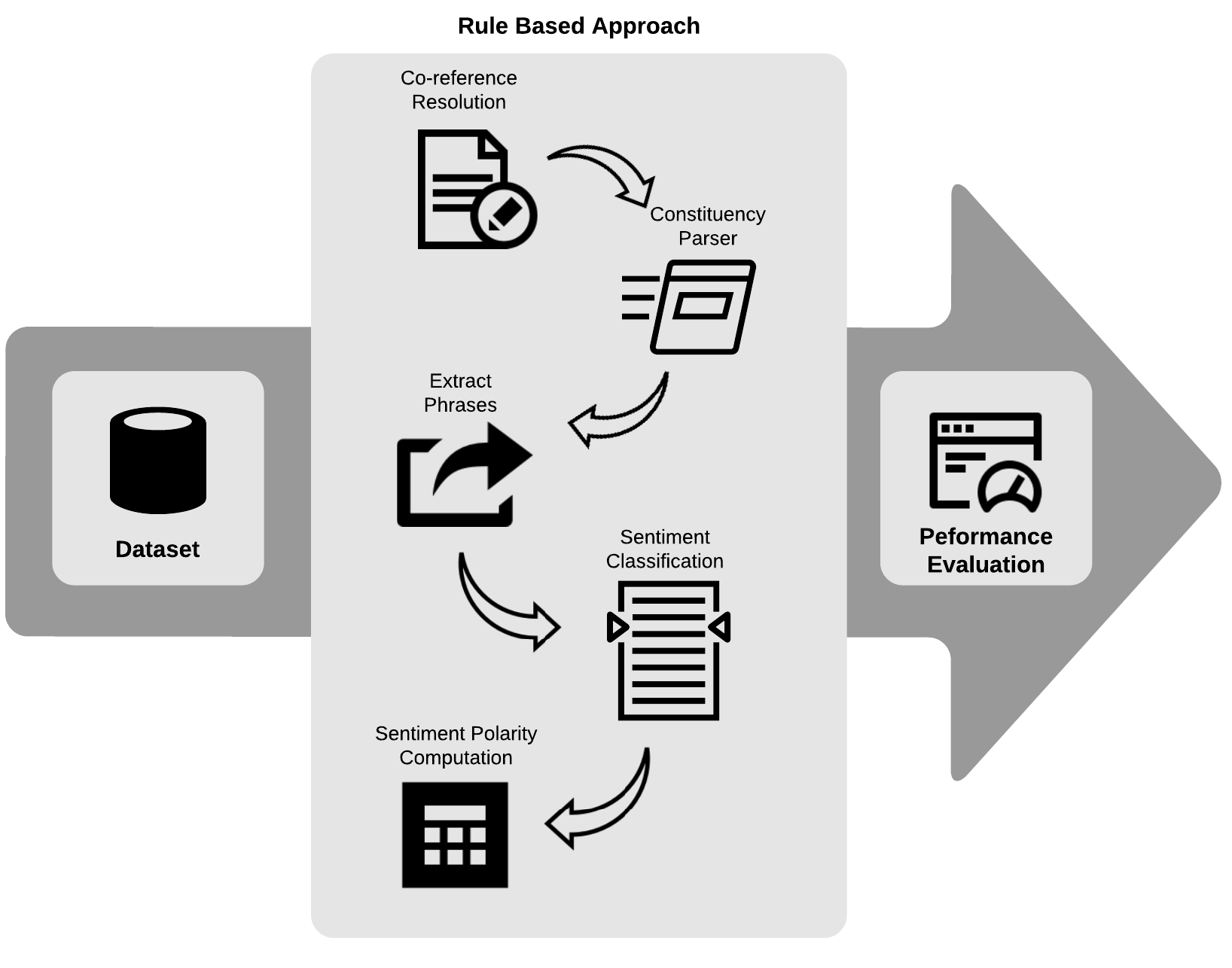}
\caption{Research Methodology}
\label{methodology}
\end{figure}

\subsection{Co-reference Resolution}

 Co-reference resolution is the function of identifying all expressions in a text that corresponds to the same real-world object \cite{clark2016improving}. In a court case, there can be some sentences which have used pronouns to mention members of a particular party. In this process, it is important to find out  sets of co-referring expressions in a sentences referring to the same aspect (party) and cluster them in order to get overall polarity value towards a particular party. Stanford Co-reference Resolution model is used due to its proven performance over many NLP tasks. 

\subsection{Generating sub sentences}

We used the constituency parser of Stanford CoreNLP \cite{manning2014stanford} for the process of generating sub sentences. Legal documents usually contain longer sentences with many subordinate sentences. When a sentence has several subordinate sentences becomes even more complex for a machine to understand \cite{ratnayaka2019shift}. Hence the sentence is split using subordinating conjunctions and extracted subordinate clauses as sub sentences. After obtaining the parse tree of the sentence using constituency parser, sub sentences are generated by splitting the parse tree by the \textit{SBAR} tags. Here \textit{SBAR} tags are used to label the subordinate conjunctions.

\subsection{Extracting phrases}

According to the phrase-structure grammar, the sentence (S) is mainly divided into two phrases (Noun Phrase-NP and  Verb Phrase-VP). NP contains the noun and its modifiers while VP consists of a main verb, helping verbs, objects, or other modifiers. It is generally known that the basic pattern of a simple sentence is subject-verb-object-adverbial. The subject is typically a noun phrase. The verb recognizes an action, occurrence or state. An object is given an action and typically being acted according to the verb \cite{thet2010aspect}.

Having regard to the above facts, we have come up with a principal rule which considers that the sentiment of the verb phrase considerably affects the object. Depending on this rule, the noun phrase and the verb phrase of sub sentences are used in sentiment allocation process. Here we used the constituency-based parse tree to extract the noun and verb phrases.

\subsection{Phrase-level Sentiment Annotator} 

For our study we used the phrase level sentiment annotator for the legal texts developed by Gamage et al. \cite{gamage2018fast} in 2018.  Their model has been developed specifically for the legal domain based on transfer learning, and it can be considered as a domain adaptation task which uses the Stanford Sentiment Annotator \cite{Socher2013RecursiveDM} as the base model. In their approach Sentiment of a phrase is classified into one of negative or non-negative classes without considering any specific party involved in the sentence. In order to their approach, the sentiment classes of our approach are also negative and non-negative. 

\subsection{Sentiment Classification}

Up to this step, the sentence is divided into sub sentences and then into phrases.Then we take the contextual sentiment score and sentiment level of each verb phrases using the phrase level annotator and assign those values to the party mentioned in the verb phrase and assign opposite values to the noun phrase (assuming one phrase can have one party member). 

In one sentence, multiple phrases can have the same party members or multiple phrases can have different party members. Therefore  we grouped each party members and their sentiment scores and sentiment levels. Using the sentiment scores averaging method, the sentiments of each party member are assigned. 
After  that  party  members  are grouped  into  main legal parties;petitioner  and  defendant. Following the same scoring method, the sentiment value of the sentences with respect to petitioner and defendant parties are assigned.

\section{Conclusion}

This paper introduces a novel approach to perform party-based sentiment analysis in legal opinion texts. Overall, this paper illustrates how phrase-level sentiment values can be effectively used to obtain the aspect-level sentiment values using NLP tools and manually created rules.
 




%
\bibliographystyle{IEEEtran}
\bibliography{content}

\end{document}